\documentclass[11pt]{article}

\usepackage[utf8]{inputenc}
\usepackage[T1]{fontenc}
\IfFileExists{lmodern.sty}{\usepackage{lmodern}}{}
\usepackage[margin=1in]{geometry}
\usepackage{amsmath,amssymb}
\usepackage{graphicx}
\usepackage{float}
\usepackage{booktabs}
\usepackage{tabularx}
\usepackage{caption}
\usepackage{subcaption}
\usepackage{authblk}
\usepackage[numbers,sort&compress]{natbib}
\usepackage[colorlinks=true,linkcolor=black,citecolor=blue,urlcolor=blue]{hyperref}
\IfFileExists{microtype.sty}{\usepackage[protrusion=true,expansion=false]{microtype}}{}

\title{\textbf{Depth-Regularized JEPA World Models Learn More\\ Transferable Representations from Real Outdoor Robot Data}}

\author[1]{Usman M. Khan}
\affil[1]{Aigen}
\date{}

\begin{document}
\maketitle

\begin{abstract}
World models, especially based on JEPA architectures, have shown strength in learning robust dynamics of various environments. While there have been great successes in simple, simulated environments, learning from visually complex real-world data remains a challenge, especially in unpredictable outdoor environments. We introduce depth as a geometric prior during training in learning more robust latent dynamics directly from robot video data and handling visual complexity. We augment LeWorldModel (LeWM) with a training-only depth regularization objective that aligns RGB embeddings with embeddings computed from paired depth camera observations, while keeping inference RGB-only. Viewed together, LeWM's isotropy-inducing latent regularizer (SIGReg) maximizes task-agnostic latent diversity while the depth term constrains how that diversity is organized, so the combined objective targets the highest-entropy representation consistent with scene geometry rather than a purely isotropic one. To satisfy this greater complexity without increasing inference time, we also add training-only overparameterization. Training an 18M-parameter model on video from a single real agricultural robot platform, we evaluate with frozen-representation visual odometry probes, predictor-based surprise detection, and multi-step latent rollout fidelity. Compared to the baseline LeWM, our method lowers visual odometry probe error by 33\%, substantially increases surprise-score separation both in-domain and on the out-of-domain TartanGround benchmark, and improves multi-step rollout fidelity under domain shift, with gains that grow with rollout horizon. Notably, we also see improvements in surprise-score separation on physics understanding that is not directly tied to 3D geometry, such as lighting and shadows. These results show that a lightweight training-time geometric prior makes a compact JEPA world model more useful and more transferable on real outdoor data with strong underlying representations, without requiring depth at test time. Our work suggests that depth as a physically grounded prior can enhance world model generalization on a variety of tasks.
\end{abstract}

\section{Introduction}
One of the key challenges in robotics is to create artificial intelligence that is general-purpose in understanding, generalizing across diverse visual environments. To this end, world models, whose objective is to predict future states based on current states, have shown promise for their ability to model environmental dynamics, a task formulation that lends itself to many different downstream tasks. One of the dominant paradigms in the world-model discourse is to leverage generative models using pixel-level reconstruction as their core task in the world-modeling objective. While these approaches are showing progress in many structured, canonical robotic domains, unstructured environments continue to present unique challenges. Outdoor robots, such as those operating in agricultural settings, must function under changing sunlight, shadows, motion blur, foliage clutter, terrain variation, and strong appearance shifts in short amounts of time. In this regime, direct pixel prediction is especially vulnerable to modeling nuisance variation rather than scene structure.

JEPA-style world models \citep{assran2023ijepa,bardes2024vjepa} offer an appealing alternative: they predict future latent representations rather than future pixels. This shifts the objective toward abstractions that are more stable than raw appearance and better matched to downstream reasoning. However, handling the visual complexity of the real world without representational collapse with this approach remains a challenge. These models are often developed and evaluated in controlled simulated domains or visually simpler indoor settings, which provide cleaner training signals. V-JEPA \citep{bardes2024vjepa} demonstrated that large-scale latent video prediction yields strong motion and appearance understanding from real video, and its successors \citep{assran2025vjepa2} added action-conditioned stages enabling zero-shot robot planning on real platforms, but these systems are resource-intensive and complex to train. LeWorldModel \citep{maes2026lewm} introduced a simplified training recipe that uses a single regularizing term to avoid collapse. Rather than relying on supervision or heuristics, LeWM presents a stable training approach based purely on the data by encouraging isotropic Gaussian feature representations. This results in stable performance even on compact models, with impressive results on a variety of simulated tasks. However, LeWM's evaluation has so far been confined to simulated benchmarks; whether the recipe survives contact with visually complex, real-world outdoor data---and what additional training-time structure helps it do so---remains untested. This is the gap we address.

We investigate a simple intervention: a training-only depth regularization loss. Since depth provides supervision related to physical 3D structure, this acts as a complementary signal alongside the diversity-maximizing SIGReg term. The two play distinct and compatible roles. SIGReg is a maximum-entropy pressure: an isotropic Gaussian is the highest-entropy latent configuration at fixed scale, which maximizes task-agnostic diversity and prevents collapse. Depth, in contrast, acts as a geometric constraint that orients that diversity toward physically meaningful directions. Their combination can be read as seeking the highest-entropy latent distribution consistent with scene geometry---a structured, geometry-informed prior rather than a purely isotropic one, which is the appropriate target once the downstream setting is known to be geometric rather than arbitrary. During training, paired RGB and stereo depth observations are passed through the same encoder--projector stack, and the RGB embedding is encouraged to align with the corresponding depth embedding via a cosine loss with stop-gradient on the depth branch. At inference, the model uses RGB alone. Intuitively, depth acts as a geometric teacher during training, nudging the JEPA representation toward scene structure that should transfer better across appearance changes and across domains.

This structured objective is, however, harder to satisfy at fixed capacity: the encoder must simultaneously produce latents that are predictable, isotropic, and geometry-aligned. Because a capacity-limited predictor can only exploit simple latent dynamics, it exerts back-pressure on the encoder to keep representations simple and predictable, competing with the geometric structure we aim to induce. We therefore introduce predictor-side overparameterization, which absorbs part of this predictive burden and frees the encoder to retain geometric structure---an intervention on the same predictor bottleneck our analysis identifies (Section~\ref{sec:discussion}). Like depth, this capacity is privileged to training: the added branches are algebraically merged into the base predictor weights after training, in the style of structural re-parameterization \citep{ding2021repvgg,vasu2023mobileone}, so the deployed model is unchanged.

We use a compact 18M-parameter LeWM variant \citep{maes2026lewm} trained on real agricultural robot data collected by the Aigen fleet. We compare a vanilla RGB model against the same architecture trained with our protocol. We focus on three questions: (i) whether depth regularization improves representation quality; (ii) whether it strengthens physically meaningful prediction behavior; and (iii) whether the resulting latent space transfers better out of domain. Across visual odometry probing, surprise evaluation, and multi-step rollout fidelity, the depth-regularized model is consistently more physics-aware and more transferable.

\paragraph{Contributions.}
\begin{itemize}
  \item We demonstrate that a compact end-to-end JEPA world model can be trained directly on real outdoor agricultural robot data, to our knowledge a setting not previously reported for this architecture family.
  \item We introduce a training-only depth alignment objective that injects geometric structure into the JEPA representation, boosting physical understanding on real data without requiring depth at inference, task labels, or a pretrained encoder.
  \item We show, under a consistent evaluation protocol spanning VO probing, surprise detection, and multi-step latent rollout evaluation, that this objective improves representation quality, predictor sharpness, and out-of-domain robustness on the TartanGround benchmark \citep{patel2025tartanground}.
\end{itemize}

\section{Related Work}
\subsection{Latent world models and JEPA-style prediction}
World models have long been framed as predictive models of environment dynamics that support planning, control, or policy learning \citep{ha2018worldmodels,hafner2019dreamer}. A major divide in this literature is between methods that reconstruct observations in pixel space \citep{hafner2019dreamer,hafner2019planet} and methods that predict in latent space. JEPA-style approaches \citep{assran2023ijepa,bardes2024vjepa,assran2025vjepa2} belong to the latter category. Rather than modeling every pixel, they encode and predict only the aspects of the world that are stable and useful for future reasoning.

This framing is especially appealing for outdoor robotics. Agricultural scenes contain many unpredictable or task-irrelevant image details---leaf texture, lighting shimmer, background clutter. A model that predicts latent structure rather than pixels is therefore better aligned with the hypothesis that useful embodied representations should focus on geometry, motion, and physical continuity.

\subsection{LeWorldModel and compact end-to-end JEPA training}
Our work builds directly on LeWorldModel (LeWM) \citep{maes2026lewm}, which shows that a compact JEPA can be trained end-to-end from raw pixels using a simple prediction objective and latent regularization (SIGReg). SIGReg solves the mode-collapse problem characteristic of many JEPA models by regularizing latents toward isotropic Gaussians, thereby guiding the model toward representations that maximize task-agnostic diversity---full-rank, non-collapsed latents that make no commitment about which directions of variation matter. LeWM provides the base architecture and training recipe used here. Our question is different: once the model is moved from benchmark-style simulated settings into real outdoor robot data, what can we do at training time to create more robust representations in the face of this enhanced visual complexity?

V-JEPA previously demonstrated promising performance on real-world video \citep{bardes2024vjepa}. However, it is a much larger model and is also more complicated to train, which constrains practical deployments in robotics. Our work is aimed at combining the general-purpose robustness SIGReg provides with a soft inductive bias that guides latents to focus on real physical structure, thereby facilitating better learning of physical dynamics.

\subsection{JEPA world models in robotics}
Recent large-scale JEPA-based systems have shown that latent predictive modeling can support robotic planning when combined with substantial pretraining or staged training pipelines \citep{assran2025vjepa2}. Those systems demonstrate that JEPA-style representations are relevant to robotics, but they operate in a different regime from the one we study here, namely that of large-scale pretraining and corresponding model size. Instead, we study whether a compact end-to-end JEPA can learn from real outdoor fleet data directly, and whether a lightweight training-time geometric prior can make that approach more effective.

\subsection{Geometry-aware supervision}
Our work is also related to methods that use privileged or geometric supervision to shape visual representations \citep{kabra2026omnivorous,vapnik2009privileged,hinton2015distilling}. Closest to our approach, \citet{kabra2026omnivorous} show that off-the-shelf encoders such as DINOv2 exhibit poor cross-modal alignment---RGB and depth embeddings of the same scene are barely more similar than those of unrelated images---and introduce a post-training adapter that aligns RGB and depth embeddings of a frozen foundation encoder, conferring robustness to input shift. We perform the analogous alignment end-to-end inside a compact world model trained from scratch, with no frozen teacher and no adapter. Depth is particularly attractive because it preserves scene structure while suppressing much of the appearance variation that dominates RGB imagery. However, in practical settings it is not always available as an input. Accordingly, we use it only as a training-time teacher that biases RGB embeddings toward a more geometry-aware latent space, by aligning their projections in embedding space.

\subsection{Inductive bias in JEPA models}
A recurring theme across JEPA world models is that the latent-prediction objective alone is underdetermined and prone to collapse, so many methods introduce some additional inductive bias to shape the representation. These biases take several forms. The most common are anti-collapse \emph{heuristics}: masked-prediction JEPAs such as I-JEPA \citep{assran2023ijepa} and V-JEPA \citep{bardes2024vjepa,assran2025vjepa2} rely on an exponential-moving-average target encoder together with stop-gradient updates, which stabilize training but do not correspond to the minimization of a well-defined objective \citep{tian2021understanding}. A second family adds \emph{auxiliary supervision or privileged signals}: PLDM \citep{sobal2025pldm} augments the latent-prediction loss with VICReg-style regularization terms; other action-conditioned world models incorporate proprioceptive inputs or action-decoder objectives \citep{assran2025vjepa2}; and in the driving setting, LAW \citep{li2024law} supervises latent prediction with future waypoint regression, while World4Drive \citep{zheng2025world4drive} relies on auxiliary pretrained encoders that require labeled data. A third family borrows inductive bias wholesale from a \emph{frozen foundation encoder}, as in DINO-WM \citep{zhou2024dinowm}, trading end-to-end learning for guaranteed non-collapse.

A distinct line builds inductive bias into the \emph{architecture} itself rather than the loss. seq-JEPA \citep{ghaemi2025seqjepa} learns architecturally segregated invariant and equivariant representations through sequential prediction over action--observation pairs, obtaining equivariance without explicit equivariance losses or dual predictors; related work enforces group-structured or homomorphic latent dynamics so that actions act as consistent transformations in latent space \citep{vanderpol2020mdp}. Others impose a \emph{domain-structured} latent, such as AD-L-JEPA's \citep{zhu2025adljepa} bird's-eye-view embedding space for LiDAR driving data. Against this backdrop, LeWM \citep{maes2026lewm} is largely a reaction against the heuristic end of the spectrum: it removes EMA, stop-gradient, pretrained encoders, and multi-term losses, replacing them with a single principled regularizer (SIGReg) that provably prevents collapse. Our work occupies a narrow position within this landscape. Rather than a heuristic, a hard architectural constraint, or a reconstruction task, we add a \emph{soft, training-only supervisory} bias---geometric alignment to paired depth---that is removed at inference and layered on top of LeWM's principled core. In the taxonomy above it is closest to the auxiliary-supervision family, but unlike waypoint or reward supervision it requires no task labels, and unlike a frozen foundation encoder it preserves fully end-to-end training: a lightweight geometric prior that complements, rather than replaces, SIGReg's anti-collapse guarantee.

\section{Method}
\subsection{Base world model}
\begin{figure}[H]
  \centering
  \includegraphics[width=\linewidth]{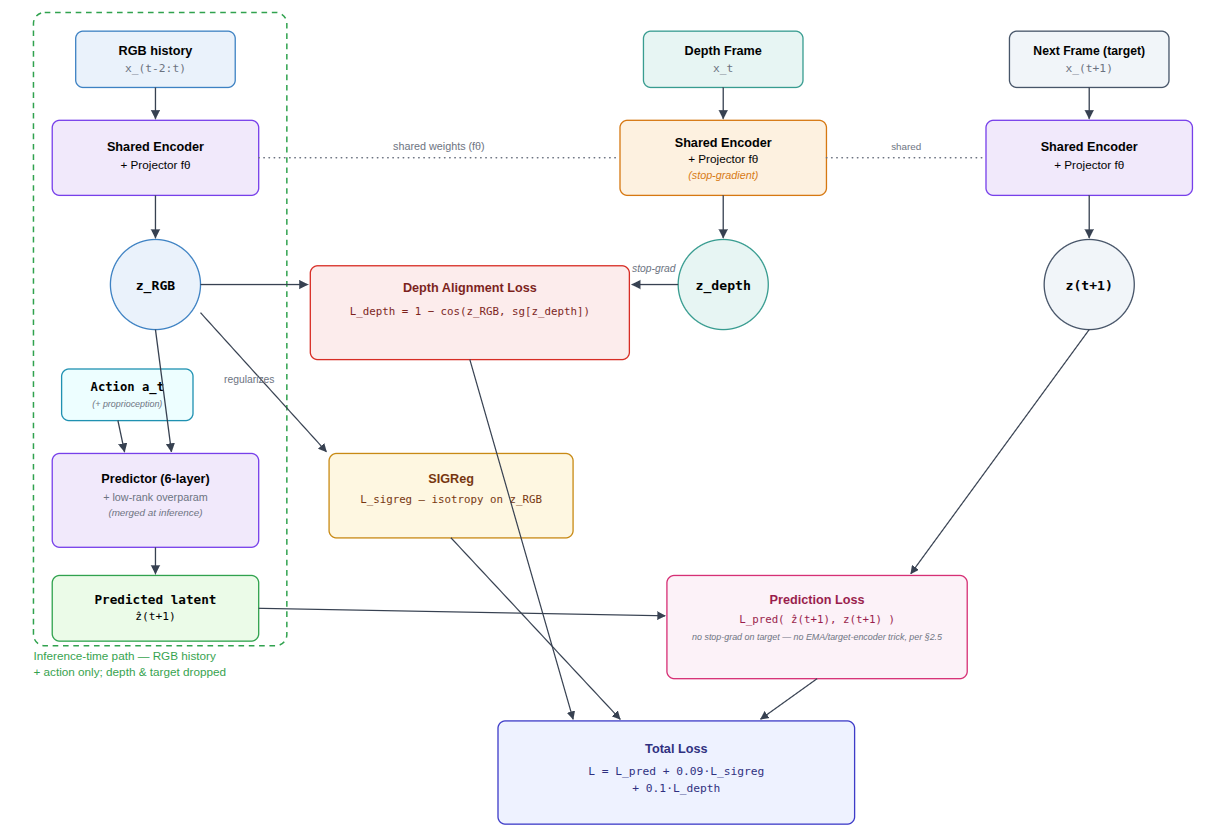}
  \caption{Method overview. RGB history is encoded by the shared encoder--projector $f_\theta$ and, together with the action, drives the predictor to produce $\hat{z}(t{+}1)$. This is compared against the true next-frame embedding $z(t{+}1)$ (same encoder, full gradient) via the prediction loss $\mathcal{L}_{\text{pred}}$. In parallel, the RGB embedding is aligned with a stop-gradient depth embedding ($\mathcal{L}_{\text{depth}}$) and regularized toward isotropy (SIGReg); all three losses combine into the total training objective. At inference (green dashed box), only the RGB history $+$ action path runs, the low-rank predictor branches are merged into the base weights, and the depth and target-frame branches are dropped entirely.}
  \label{fig:method}
\end{figure}
We build on LeWorldModel (LeWM), a compact JEPA world model that predicts future latent embeddings from image history, actions, and proprioception. In the real-data configuration used here, the model consumes $224 \times 224$ RGB inputs, uses history size 3, predicts 1 step ahead, has embedding dimension 192, and uses a ViT-Tiny encoder with a 6-layer predictor. Both variants (vanilla and depth-regularized) are trained for 100 epochs with identical optimizer, schedule, and batch configuration. Repeated training runs yielded consistent results; we report a representative run.

We additionally include predictor-side overparameterization during training to augment the model's expressive power, compensating for the additional loss term. In the current implementation, the base 6-layer LeWM predictor is augmented during training with rank-16 low-rank RepLinear branches on each predictor MLP up-projection and down-projection. Because these branches are linear, they are algebraically merged into the base weights after training \citep{ding2021repvgg,vasu2023mobileone}; the inference-time predictor is architecturally identical to the baseline and incurs no additional cost. Capacity, like the depth signal, is thus privileged to training and free at test time. The prediction loss couples the encoder and predictor, so a capacity-limited predictor, which is able to exploit only simple latent dynamics, can pressure the encoder to keep its latents simple and predictable, competing with the isotropic, geometry-aligned structure that SIGReg and the depth term induce.

\subsection{Training-only depth regularization}
The baseline is trained with the standard latent prediction loss plus the SIGReg latent regularizer \citep{maes2026lewm}. The depth-regularized variant adds a depth alignment term:
\begin{equation}
\mathcal{L} = \mathcal{L}_{\text{pred}} + 0.09\,\mathcal{L}_{\text{sigreg}} + 0.1\,\mathcal{L}_{\text{depth}}.
\end{equation}
For paired RGB and depth frames (supplied as grayscale replicated to 3 channels), the same encoder--projector network processes both modalities. The depth branch is stop-gradient; the RGB branch is optimized to maximize cosine similarity to the depth embedding:
\begin{equation}
\mathcal{L}_{\text{depth}} = 1 - \cos\!\left(f_\theta(x^{\text{RGB}}),\ \operatorname{sg}\big[f_\theta(x^{\text{depth}})\big]\right).
\end{equation}
The SIGReg weight follows the value used in LeWM. The depth weight of 0.1 was chosen heuristically to be of comparable magnitude to the SIGReg term. Depth frames are used only for the shared encoder/projector alignment branch during training, with stop-gradient on the depth side; no depth input is required at inference.

The depth term and SIGReg play complementary roles in shaping the latent distribution. SIGReg maximizes entropy, pushing the aggregate embedding toward an isotropic Gaussian; the depth term imposes a geometric constraint on how that variance is organized. The combined objective can therefore be read as targeting the maximum-entropy latent representation consistent with scene geometry. The resulting objective moves toward a controlled, geometry-informed anisotropy rather than pure isotropy. This also connects to the task-conditioning view of the regularizer: isotropy is the optimal latent target under ignorance of the downstream task, whereas an embodied, geometric setting warrants deviating toward the structure the task actually requires.

\subsection{Data and evaluation setting}
\begin{figure}[t]
  \centering
  \begin{subfigure}[b]{0.48\linewidth}
    \includegraphics[width=\linewidth]{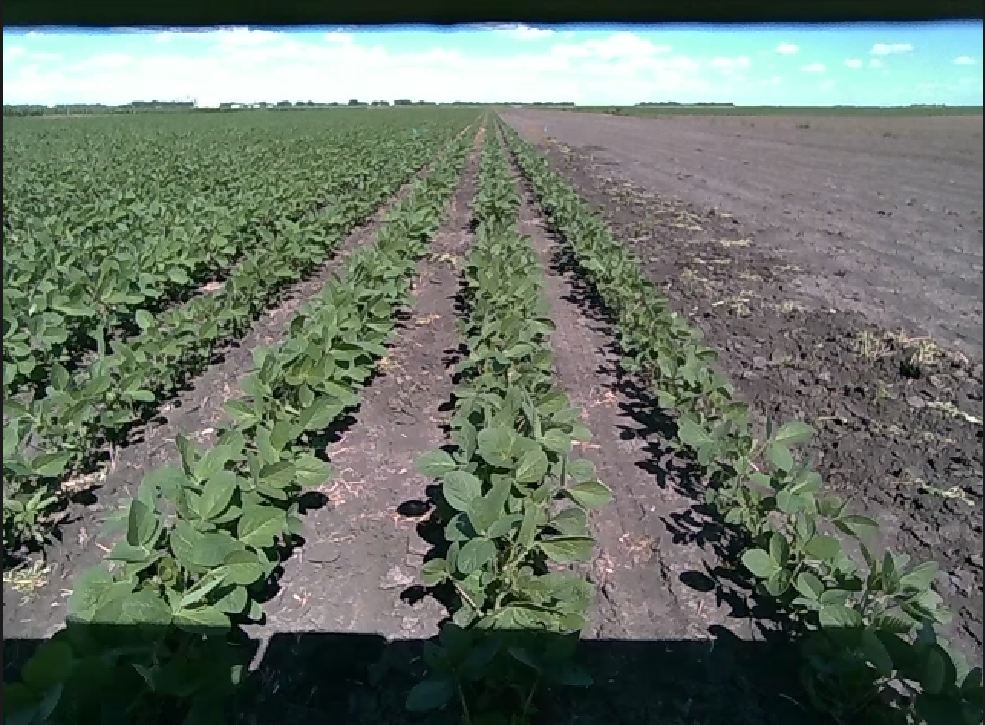}
    \caption{RGB frame.}
  \end{subfigure}\hfill
  \begin{subfigure}[b]{0.48\linewidth}
    \includegraphics[width=\linewidth]{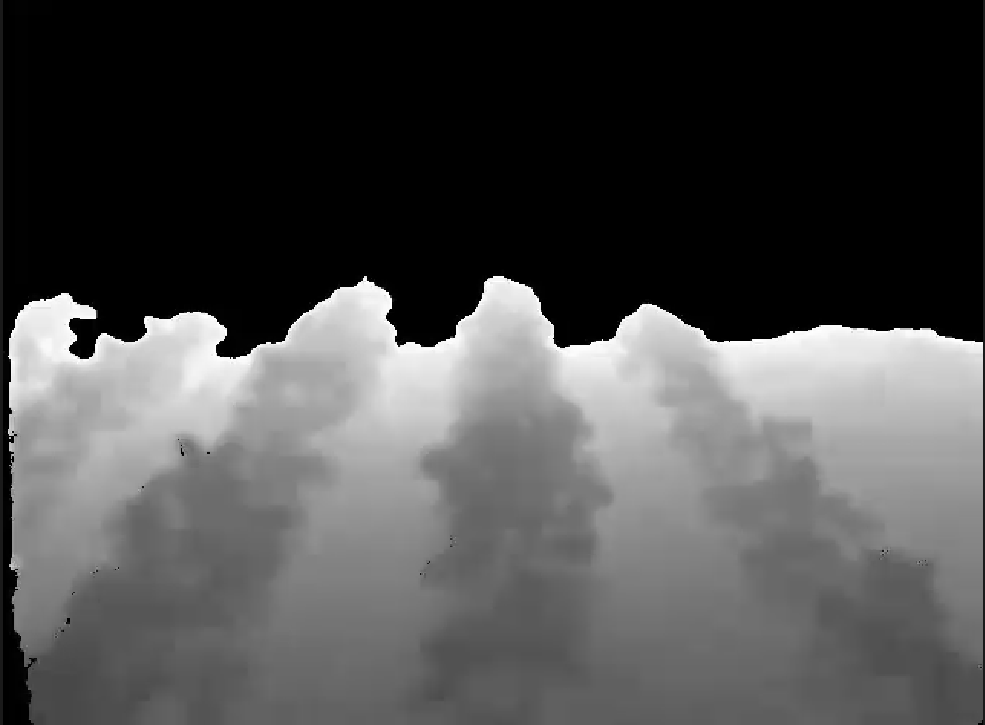}
    \caption{Depth map (1\,m cutoff).}
  \end{subfigure}
  \caption{Example data sample collected from the Aigen Element robot. (a) An RGB image from a representative scene. (b) Its corresponding depth map with a 1-meter cutoff.}
  \label{fig:data}
\end{figure}

Training uses approximately 31K frames across 85 episodes of real agricultural robot data collected on Aigen's Element platform. The data consists of driving video from a forward-facing navigational stereo camera on the robot across diverse environmental conditions. Depth frames are truncated at a distance of 1 meter. Validation uses 3{,}587 held-out frames from 9 episodes. For out-of-domain evaluation, we use TartanGround \citep{patel2025tartanground}, a ground-robot benchmark from which we draw 15{,}608 frames across 6 episodes spanning multiple environments. This OOD setting changes the robot platform, the environment, and the sim-vs-real gap simultaneously.

We evaluate the learned models on three axes:
\begin{enumerate}
  \item A \textbf{frozen-embedding visual odometry linear probe} that regresses ego-motion from the model's embeddings.
  \item \textbf{Predictor-based surprise detection} under synthetic perturbations of the observation stream.
  \item \textbf{Multi-step latent rollout fidelity} under ground-truth action sequences.
\end{enumerate}
We additionally include a frozen DINOv2 \citep{oquab2023dinov2} encoder as a non-LeWM reference point on the VO probe, to contextualize results against a widely used single-frame visual representation.

\subsection{Surprise evaluation protocol}
The surprise evaluation tests whether the world model has learned physically meaningful expectations about how the scene should evolve under motion. We follow a predictor-based setup: given an observation and action, the model predicts the next latent embedding, and surprise is measured as the squared error between that prediction and the actual next embedding. Higher surprise should correspond to transitions that violate the model's learned expectations.

To probe this behavior, we construct short clips containing either plausible transitions or synthetically perturbed ones. The perturbations are chosen to break different aspects of real-world consistency: \textbf{teleportation} introduces an abrupt spatial discontinuity; \textbf{temporal reversal} violates normal motion direction; \textbf{brightness jump} creates an implausible illumination change between frames; and \textbf{color swap} disrupts appearance consistency while preserving coarse layout. These perturbations test whether the model has learned that outdoor scenes respond coherently to ego-motion, and that lighting and appearance should evolve with some temporal continuity rather than arbitrarily jumping frame to frame.

We report both classification accuracy and the separation between surprise scores on perturbed versus plausible clips. Surprise is computed from the predictor error between the predicted next latent and the actual next latent under the same action context; separation is the mean perturbed surprise minus the mean plausible surprise. In practice, accuracy can saturate on easy perturbations, so separation is often the more informative metric: it indicates the margin between physically plausible and implausible transitions, which is the desired behavior for a world model meant to encode motion-conditioned structure rather than memorize appearance.

\subsection{Rollout evaluation protocol}
The rollout evaluation measures how faithfully the predictor tracks real trajectories over multiple steps, isolating the quality of the learned dynamics from any action-selection procedure. Starting from an encoded observation history, we unroll the predictor autoregressively over the ground-truth action sequence for a horizon $h$ and compare the predicted latents against the encodings of the actually observed future frames. We report two cosine-similarity metrics: \textbf{goal similarity}, between the predicted latent at the final step and the encoding of the true frame at $t+h$, and \textbf{trajectory similarity}, the mean similarity across all intermediate steps. We evaluate at $h=10$ and $h=20$ on both Aigen validation and TartanGround. This evaluates the model's ability to predict future world dynamics given an input state sequence, which represents the central world-modeling objective.

\section{Results}
\begin{table}[t]
  \centering
  \caption{Main results, vanilla LeWM vs.\ depth-regularized (ours). Surprise accuracy is omitted here---per Section~\ref{sec:surprise}, both models saturate on accuracy, so separation is the more informative signal. Best value in each row in \textbf{bold}.}
  \label{tab:main}
  \small
  \begin{tabularx}{\linewidth}{Xccc}
    \toprule
    Metric & Vanilla LeWM & Depth-Reg (Ours) & Frozen-DINOv2 ref. \\
    \midrule
    VO probe MSE $\downarrow$ & 0.0022 & \textbf{0.0015} & 0.0029 \\
    Surprise separation --- Aigen-val $\uparrow$ & 0.067 & \textbf{0.120} & 0.039 \\
    Surprise separation --- TartanGround (OOD) $\uparrow$ & 0.035 & \textbf{0.099} & 0.030 \\
    Rollout goal sim.\ ($h{=}10$) --- Aigen-val $\uparrow$ & 0.903 & 0.890 & \textbf{0.951} \\
    Rollout traj.\ sim.\ ($h{=}10$) --- Aigen-val $\uparrow$ & 0.967 & 0.956 & \textbf{0.971} \\
    Rollout goal sim.\ ($h{=}10$) --- TartanGround (OOD) $\uparrow$ & 0.716 & \textbf{0.745} & 0.656 \\
    Rollout traj.\ sim.\ ($h{=}10$) --- TartanGround (OOD) $\uparrow$ & 0.861 & \textbf{0.873} & 0.812 \\
    Rollout goal sim.\ ($h{=}20$) --- TartanGround (OOD) $\uparrow$ & 0.458 & \textbf{0.537} & 0.533 \\
    Rollout traj.\ sim.\ ($h{=}20$) --- TartanGround (OOD) $\uparrow$ & 0.724 & \textbf{0.767} & 0.695 \\
    \bottomrule
  \end{tabularx}
\end{table}

\subsection{Improved representation quality}
A linear probe from frozen embeddings to ego-motion shows a clear gain from depth regularization on the visual odometry task. Validation MSE drops from \textbf{0.0022} for the vanilla LeWM to \textbf{0.0015} for the depth-regularized model---a 33\% relative reduction. A frozen DINOv2 encoder, included as a non-LeWM reference, reaches 0.0029, which is worse than either LeWM variant. That a compact world model trained from scratch on ${\sim}31$K in-domain frames yields more linearly decodable ego-motion than a large pretrained single-frame representation underscores the value of in-domain temporal training, and depth regularization widens the gap further. This is the cleanest evidence that the latent representation captures motion-relevant geometric structure more faithfully when depth is used as a training signal.

\subsection{Increased surprise separation, especially out of domain}
\label{sec:surprise}
We also examine whether the predictor assigns higher surprise to physically implausible events than to plausible sequences. The perturbations probe expectations about spatial continuity, motion-consistent temporal order, and the fact that outdoor illumination and appearance should not change arbitrarily from one step to the next.

Accuracy is often saturated on easy perturbations, so the more revealing signal is the separation between perturbed and plausible surprise scores. On the held-out Aigen validation set, both models obtain the same overall detection accuracy of \textbf{78.7\%}, but mean surprise separation increases from \textbf{0.067} to \textbf{0.120} with depth regularization. On TartanGround, overall accuracy rises from \textbf{87.5\%} to \textbf{89.2\%}, and separation increases from \textbf{0.035} to \textbf{0.099}---a nearly $3\times$ improvement in margin under full domain shift. Among the geometric perturbations, the largest gains appear on teleportation, indicating that the depth-regularized latent space yields a predictor with stronger confidence margins when scene geometry is violated. Gains on the appearance-based perturbations are treated separately in the next subsection.

Beyond anomaly detection, this suggests that the model has learned a better prior over how the world should respond to ego-motion.

\subsection{Depth improves even non-geometric physical understanding, such as lighting}
A priori, one would expect a geometric prior to help most on geometric violations. The more surprising finding is that depth regularization also sharpens surprise separation on perturbations with no direct 3D content: brightness jumps and color swaps. On Aigen validation, separation on brightness-jump perturbations increases from 0.114 to 0.247, and color-swap separation increases from 0.092 to 0.135. On TartanGround, the corresponding separations increase from 0.032 to 0.124 for brightness jumps and from 0.019 to 0.036 for color swaps.

One interpretation is that grounding the latent in scene structure gives the model a scaffold against which appearance must remain consistent: once the representation encodes 3D layout, illumination that changes without a corresponding geometric cause becomes easier to flag as implausible. Under this reading, the depth prior does not merely improve geometric readout; it also induces physically structured expectations about appearance. A weaker alternative explanation is that any well-chosen auxiliary objective tightens the latent space and generically sharpens prediction-error margins; distinguishing these accounts would require a non-geometric auxiliary control, and a split of lighting perturbations into geometry-inconsistent versus globally uniform shifts.

\subsection{Improved OOD generalization}
In domain, the depth-regularized model tracks slightly below the vanilla baseline (goal similarity 0.890 vs.\ 0.903 at $h=10$). Under full domain shift the ordering reverses: on TartanGround the depth model is better at $h=10$ (0.745 vs.\ 0.716 goal similarity) and the advantage widens substantially at $h=20$ (0.537 vs.\ 0.458)---the gain grows with rollout horizon (Table~\ref{tab:main}). This is the pattern expected from a model that trades some in-domain specialization for a more transferable latent geometry, and it indicates that the depth-shaped dynamics are more robust to compounding error under domain shift. The frozen-DINOv2 reference obtains the highest in-domain rollout similarity (0.951) while being the weakest representation on the VO probe and surprise separation, a caution that rollout similarity partly rewards smooth or slowly-varying latents; cross-model rollout comparisons are therefore best read alongside the representation metrics, and the most meaningful comparisons are within-model, across domains and horizons.

\clearpage
\subsection{Depth aligns cross-domain latent geometry}
\begin{figure}[H]
  \centering
  \includegraphics[width=\linewidth]{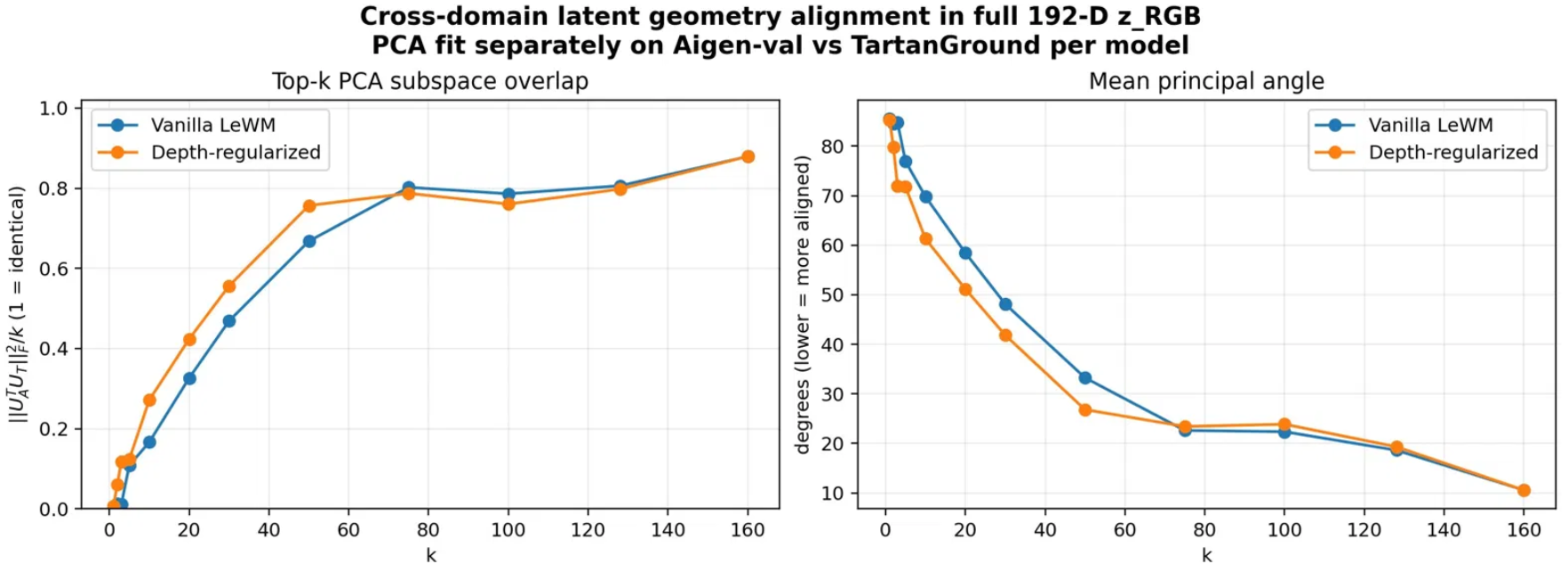}

  \vspace{1em}

  \includegraphics[width=\linewidth]{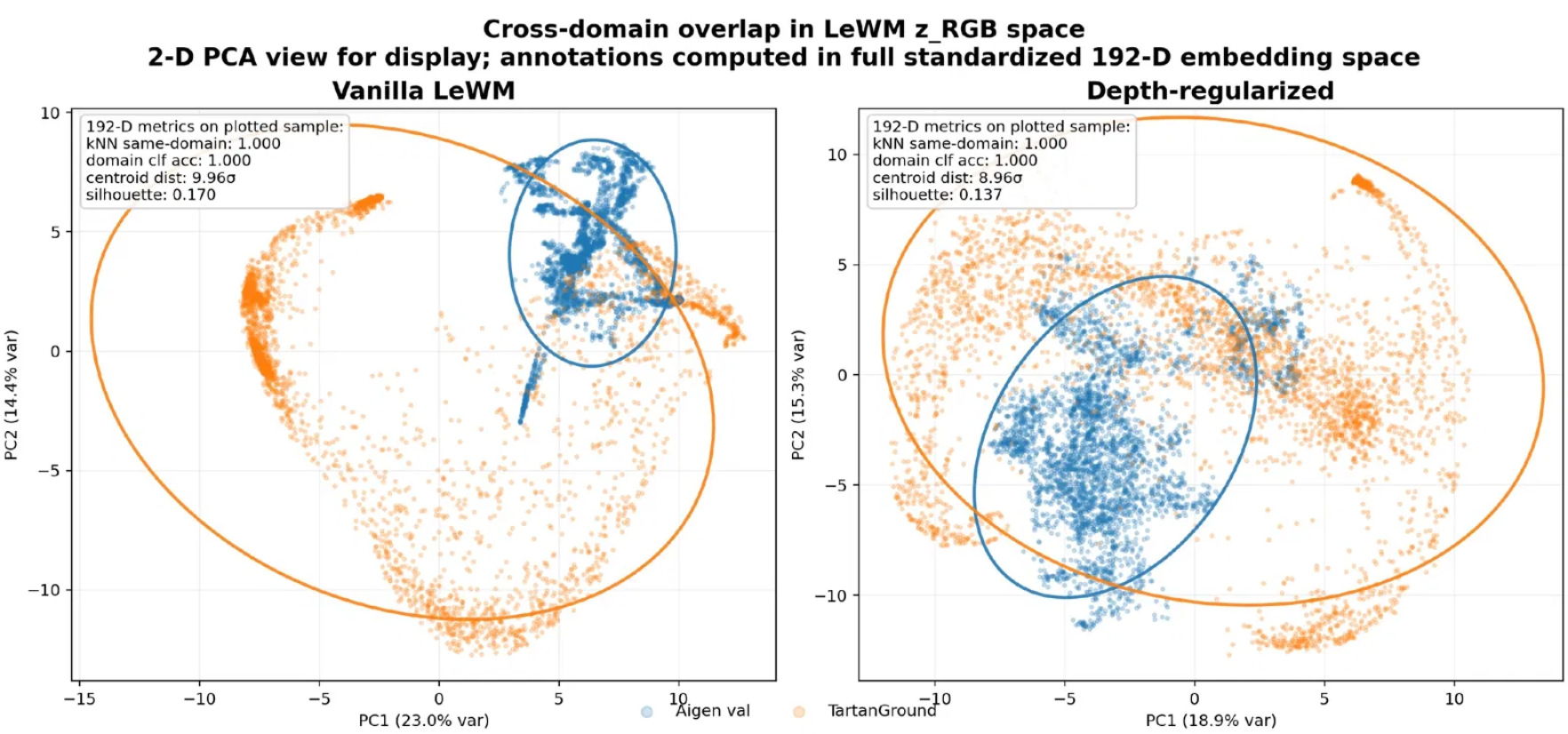}
  \caption{Cross-domain latent geometry. \textbf{Top:} principal-subspace overlap and mean principal angle vs.\ $k$ (PCA fit per domain per model, full 192-D space). The depth-regularized model is more cross-domain aligned across the mid-rank band $k \approx 5\text{--}50$; large-$k$ overlap is dimensionally forced (floor $(2k-d)/k$), so the informative regime is $k \le 96$. \textbf{Bottom:} 2-D PCA projections of Aigen validation (blue) and TartanGround (orange) embeddings ($2\sigma$ ellipses); annotated metrics are computed in the full 192-D space, not the projection. Both domains are perfectly separable under kNN and a linear classifier, as expected for real-vs-sim data; the informative comparison is the degree of separation, which decreases with depth regularization (centroid $9.96\sigma \rightarrow 8.96\sigma$; silhouette $0.170 \rightarrow 0.137$). The 2-D view understates full-space separability.}
  \label{fig:geometry}
\end{figure}

Figure~\ref{fig:geometry} offers a complementary geometric view of the transfer results. In the full 192-d embedding space, both models separate the two domains perfectly under kNN and a linear classifier. This is unsurprising, given the abundant low-level signatures available to separate the two domains. The analysis does, however, reveal an interesting contrast between the two approaches. First, the degree of separation decreases with depth regularization (Figure~\ref{fig:geometry}, bottom). Additionally, the internal structure of the two domains becomes more shared: fitting PCA separately per domain and comparing the top-$k$ principal subspaces, the depth model shows consistently higher subspace overlap and lower mean principal angles throughout the mid-rank band $k \approx 5\text{--}50$ (Figure~\ref{fig:geometry}, top). The very top components remain domain-specific in both models, as expected for directions dominated by global appearance statistics, so the alignment gain is concentrated precisely in the non-trivial mid-rank regime where task-relevant structure resides. This is direct evidence that depth regularization yields a more transferable latent geometry, complementing the behavioral OOD results.

\section{Discussion}
\label{sec:discussion}
In visually messy outdoor data, JEPA-style latent prediction is a good modeling abstraction because it avoids direct pressure to reconstruct every appearance detail. While recent work has shown remarkable performance without inductive bias, the representations in these scenarios still benefit from an additional geometric prior. Training-time depth regularization provides that prior cheaply, using paired depth only during learning and not at deployment. Our protocol also includes overparameterization at training time; this is intended to expand the predictor's capacity to exploit the enhanced signal rather than to serve as an independent source of improvement, and it is merged away at inference so it adds no deployment cost. Isolating its contribution with a dedicated capacity ablation is left to future work.

Leveraging depth as a geometric teacher as described in our work boosts the metrics most directly tied to useful representation structure: ego-motion decoding, surprise-score margin, and robustness under cross-domain transfer. These properties are strongly relevant to the objective of a generalist world model intended to support downstream planning or adaptation in changing real-world conditions. The fact that even tasks that do not directly use 3D structure saw gains suggests that moving the training objective from pure isotropy into one that additionally incorporates real physical structure creates an overall more robust latent space. This motivates future investigation into how relatively easily obtainable priors like depth can improve world models' ability to learn environmental dynamics.

Trends in modern deep learning have been moving away from inductive bias and more toward pure scaling. The protocol described in this work demonstrates performance gains purely through training-time techniques, avoiding any additional overhead at inference time. This shows that such an approach remains valuable, especially in regimes where there is a desire to keep models compact.

\section{Limitations}
This study serves as a proof of concept for the discussed methods, but has several important limitations:
\begin{itemize}
  \item \textbf{Offline, not closed-loop.} The rollout results are offline latent metrics, not closed-loop robot performance.
  \item \textbf{Planning evaluation.} We focus here on the model's ability to roll out predictions given a ground-truth action sequence. Extending this into a working robotic planner is future work; our preliminary experiments with sampling-based latent planning were inconclusive across models because the latent-similarity planning objective has a different dynamic range in each model's latent space.
  \item \textbf{Scale and data scope.} Training uses a relatively small dataset (${\sim}31$K frames) from a single robot platform in a single broad domain (outdoor row-crop agriculture). Whether the benefits of depth regularization persist at larger data and model scales is untested.
  \item \textbf{Narrow baseline comparison.} The main head-to-head is between two LeWM variants. A DINOv2 probe is included for context, but we do not compare against larger pretrained world models in this paper. A future version may include a more thorough evaluation.
\end{itemize}

\section{Conclusion}
Compact JEPA world models can learn useful latent dynamics directly from real outdoor agricultural robot data. Adding a simple training-only depth alignment loss improves geometric structure in the learned representation, sharpens the predictor's expectations, and improves multi-step rollout fidelity out of domain---with gains that grow with horizon---without requiring depth at inference. Depth regularization is therefore a practical way to inject privileged geometric supervision into RGB world models intended for real deployment: cheap at training time, free at test time.

\bibliographystyle{plainnat}
\bibliography{references}

\end{document}